\documentclass[10pt]{wlscirep}
\title{Towards Automated ICD Coding Using Deep Learning}

\author[1,2]{Haoran Shi\thanks{This work was done when the first author was an intern at Petuum Inc.}}
\author[1]{Pengtao Xie}
\author[1]{Zhiting Hu}
\author[2]{Ming Zhang}
\author[1]{Eric P. Xing}

\affil[1]{Petuum Inc, Pittsburgh, USA}
\affil[2]{Department of Computer Science and Technology, Peking University, Beijing, China}

\usepackage{bm}
\usepackage{subfigure}
\usepackage{color}
\usepackage{multirow}
\usepackage{bigstrut}
\usepackage{diagbox}

\begin{abstract}
International Classification of Diseases(ICD) is an authoritative health care classification system of different diseases and conditions for clinical and management purposes. Considering the complicated and dedicated process to assign correct codes to each patient admission based on overall diagnosis, we propose a hierarchical deep learning model with attention mechanism which can automatically assign ICD diagnostic codes given written diagnosis. We utilize character-aware neural language models to generate hidden representations of written diagnosis descriptions and ICD codes, and design an attention mechanism to address the mismatch between the numbers of descriptions and corresponding codes. Our experimental results show the strong potential of automated ICD coding from diagnosis descriptions. Our best model achieves 0.53 and 0.90 of F1 score and area under curve of receiver operating characteristic respectively. The result outperforms those achieved using character-unaware encoding method or without attention mechanism. It indicates that our proposed deep learning model can code automatically in a reasonable way and provide a framework for computer-auxiliary ICD coding.
\end{abstract}
\begin{document}
\flushbottom
\maketitle
\thispagestyle{empty}
\section*{Introduction}
The International Classification of Diseases (ICD) is a health care classification system maintained by the World Health Organization\cite{world1978international}, which provides a hierarchy of diagnostic codes of diseases, disorders, injuries, signs, symptoms, etc. It is widely used for reporting diseases and health conditions, assisting in medical reimbursement decisions, collecting morbidity and mortality statistics, to name a few. 

While ICD codes are important for making clinical and financial decisions, medical coding -- which assigns proper ICD codes to a patient admission -- is time-consuming, error-prone and expensive. Medical coders review the diagnosis descriptions written by physicians in the form of textual phrases and sentences and (if necessary) other information in the electronic medical record of a clinical episode, then manually attribute the appropriate ICD codes by following the coding guidelines\cite{o2005measuring}. Several types of errors frequently occur. First, when writing diagnosis descriptions, physicians often utilize abbreviations and synonyms, which causes ambiguity and imprecision when the coders are matching ICD codes to those labels\cite{sheppard2008ambiguous}. Second, in many cases, several diagnosis descriptions are closely related and should be combined into a single combination ICD code. However, unexperienced coders may code each disease separately. Such errors are called {\it unbundling}. Third, the ICD codes are organized in a hierarchical structure where the top-level codes represent generic disease categories and the bottom-level codes represent more specific diseases. A miscoding happens when the coder matches the diagnosis description to an overly generic code instead of a more specific code. The cost incurred by coding errors and the financial investment spent on improving coding quality are estimated to be \$25 billion per year in the US\cite{lang2007consultant, farkas2008automatic}. 

To reduce coding errors and cost, we aim at building an ICD coding machine which automatically and accurately translates the free-text diagnosis descriptions into ICD codes. To achieve this goal, several technical challenges need to be addressed. First, the diagnosis descriptions written by physicians and the textual descriptions of ICD codes are written in quite different styles even if they refer to the same disease. In particular, the definitions of ICD code are formally and precisely worded, while diagnosis descriptions are usually written in an informal and ungrammatical way, with telegraphic phrases, abbreviations, and typos. Second, as stated earlier, there does not necessarily exist a one-to-one mapping between diagnosis descriptions and ICD codes, and human coders should consider the overall health condition when assigning codes. In many cases, two closely related diagnosis descriptions need to be mapped into a single combination ICD code. On the other hand, physicians may write two health conditions into one diagnosis description which should be mapped onto two ICD codes under such circumstances.   

\section*{Contributions}
We present a deep learning approach to automatically perform ICD coding given the diagnosis descriptions. Specifically, we propose a hierarchical neural network model which is able to capture the latent semantics of ICD definitions and diagnosis descriptions, despite their significant difference in writing style. Attention mechanism is designed to address the mismatch between diagnosis description number and assigned code number. We train the model on 8,066 hospital admissions, tune hyper-parameters on 1,728 admissions, and evaluate the performance on a held-out test set of 1,729 hospital admissions. We demonstrate that our coding machine can accurately assign ICD codes.

\section*{Related work}
The accuracy and efficiency of manual ICD coding has always been a concern of clinical practice. KJ O'malley \emph{et al.} has summarized the complete workflow of assigning ICD codes manually\cite{o2005measuring}, which is a dedicated procedure and is prone to errors. To avoid the massive human labour to code, scientists have proposed some automatic or semi-automatic ICD classification system, especially from narrative clinical notes for better health care practice\cite{spyns1996natural, hearst1999untangling, zeng2006extracting, ananiadou2006text, pestian2007shared, meystre2008extracting, koopman2015automatic}. But the experimental dataset was generally small and domain specific. For example, the shared task involved assigning ICD-9 codes to 1954 radiology records has attracted a lot of attention\cite{pestian2007shared}, and In 2015 Koopman \emph{et al.} propose a classification system for identifying different types of cancers for death certificates based on ICD classification system \cite{koopman2015automatic}. Contrary to these experiments, the dataset in our experiment is much larger and contains various domains of clinical practice.

There also has been some trials to assign ICD codes utilizing the document of discharge summary\cite{larkey1996combining, franz2000automated, perotte2013diagnosis}. Leah S. Larkey and W. Bruce Croft have trained three statistical classifiers with many human-tuned parameters and an ensemble model to give candidate ICD labels, more focused on principal diagnostic code(the most significant diagnostic code), to each discharge summary document\cite{larkey1996combining}. Besides, Franz \emph{et al.} compares three coding methods given discharge diagnosis, but their object is to assign just one diagnostic code to each diagnosis description\cite{franz2000automated}. All of them utilize the full-text document of discharge summary, thus suffer from the complicated preprocessing of the noisy text. To build a more practical ICD coding machine, we formulate our coding task as a general multi-label classification problem on diagnosis descriptions, without many parameters to tune or restricting the number of assigned codes for each patient record.

\section*{Methods}

\subsection*{Dataset and preprocessing}
We perform the study on the publicly available MIMIC-III dataset\cite{johnson2016mimic}, which contains de-identified and comprehensive electronic medical records of 58,976 patient visits in the Beth Israel Deaconess Medical Center from 2001 to 2012. The patient visit record usually has a clinical note called discharge summary, which contains multiple sections of information, such as `discharge diagnosis', `past medical history', `admission medications', and `chief complaint'.  The diagnosis descriptions are usually included in the `discharge diagnosis' and `final diagnosis' sections\cite{prakash2017condensed}. We use a variety of standard text pre-processing techniques such as regular expression matching and tokenization to turn the noisy and irregular raw note texts in these sections into clean diagnosis descriptions. Each resulting label is a short phrase or a sentence, articulating one disease or condition. Patient visits that contain no extracted diagnosis descriptions are discarded. 

Each patient visit has a list of ICD codes given by the medical coders. These codes are documented in structured tables. The entire dataset contains 6,984 unique codes, each of which has a textual description, describing a disease, symptom, or condition. Many codes are only assigned to a few patient visits. Due to the sparsity of data, it is very difficult to train an accurate coding model for all of them. Instead, we choose 50 most frequent codes to carry out the study while noting that our model can readily be extended to more codes as long as sufficient training data is available. The frequency of one code is measured as the number of patient visits that the code is assigned to. 

Table~\ref{tab:example} shows a sample of admission record in the raw dataset and extracted diagnosis descriptions. The `HADMID' is used by the MIMIC-III to denote each hospital admission identically. We omitted irrelevant sections in the original texts of discharge summary, like `discharge disposition' and `physical examination'. The extracted diagnosis descriptions given by physicians are in enumeration style. Notice that there is an extra newline in the third written diagnosis description and it's removed after extraction. The number of diagnosis descriptions is not equal to the number of assigned diagnostic codes.
\begin{table}[!htbp]
  \centering
    \begin{tabular}{|l|p{30.30em}|}
    \hline
    HADMID & \multicolumn{1}{l|}{189797} \\
    \hline
    Original Texts of Discharge Summary & ...\newline{}DISCHARGE DIAGNOSIS:\newline{}1.  Prematurity at 35 4/7 weeks gestation\newline{}2.  Twin number two of twin gestation\newline{}3.  Respiratory distress secondary to transient tachypnea of\newline{}the newborn\newline{}4.  Suspicion for sepsis ruled out\newline{}...\\
    \hline
    Extracted Diagnosis Descriptions & 1. Prematurity at 35 4/7 weeks gestation\newline{}2. Twin number two of twin gestation\newline{}3. Respiratory distress secondary to transient tachypnea of the newborn\newline{}4. Suspicion for sepsis ruled out \\
    \hline
    Assigned ICD Diagnostic Codes &  `V3100', `76518', `7756', `7706', `V290', `V053' \\
    \hline
    \end{tabular}%
    \caption{One admission sample from MIMIC-III dataset.}
  \label{tab:example}%
\end{table}%

\begin{figure}[!ht]
  \centering
  \subfigure[Distribution of diagnosis description count]{
    \label{fig:dist:exp:a}
    \includegraphics[width=0.3\linewidth]{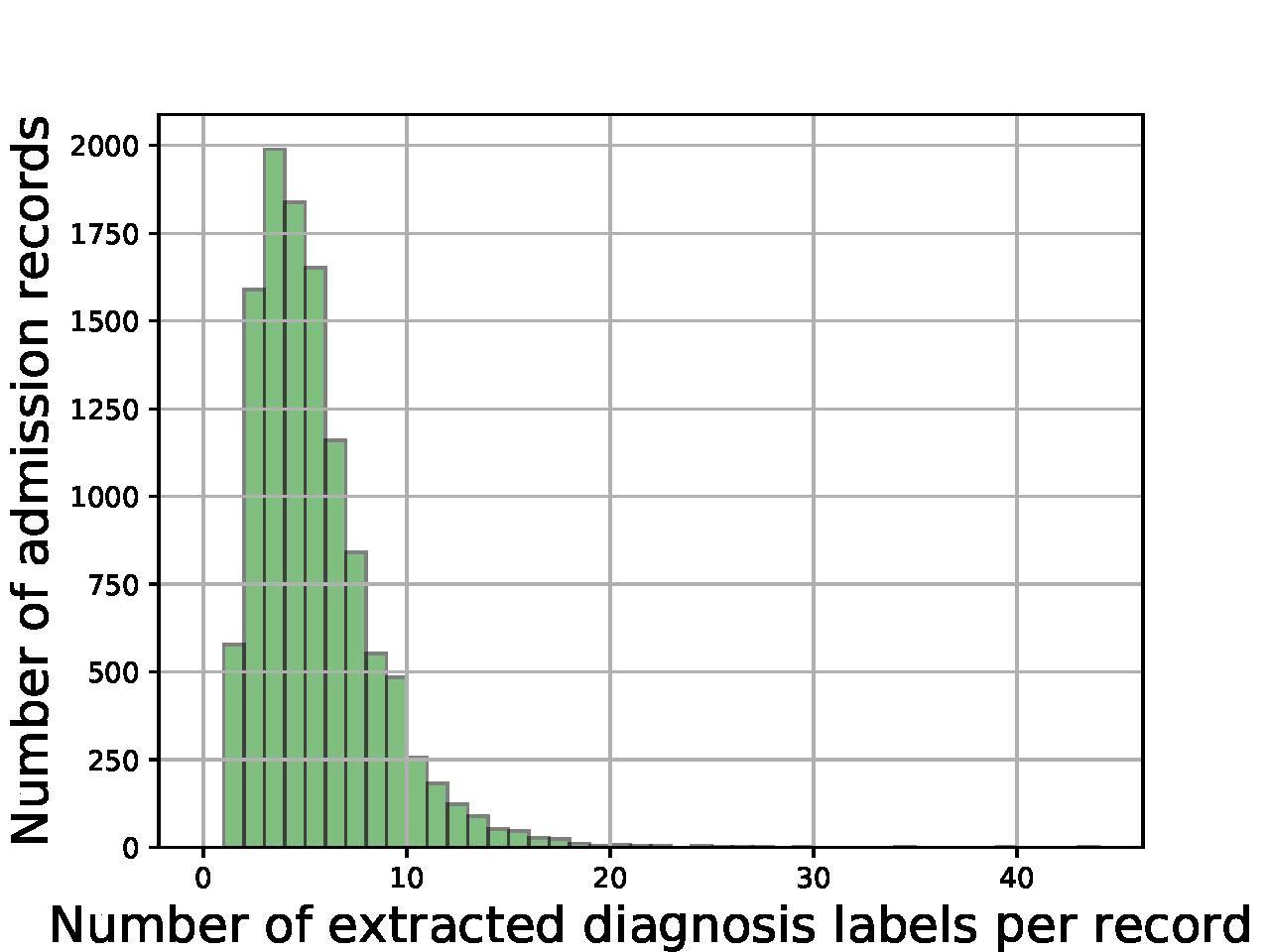}}
  \subfigure[Distribution of ICD code frequency]{
    \label{fig:dist:exp:b}
    \includegraphics[width=0.3\linewidth]{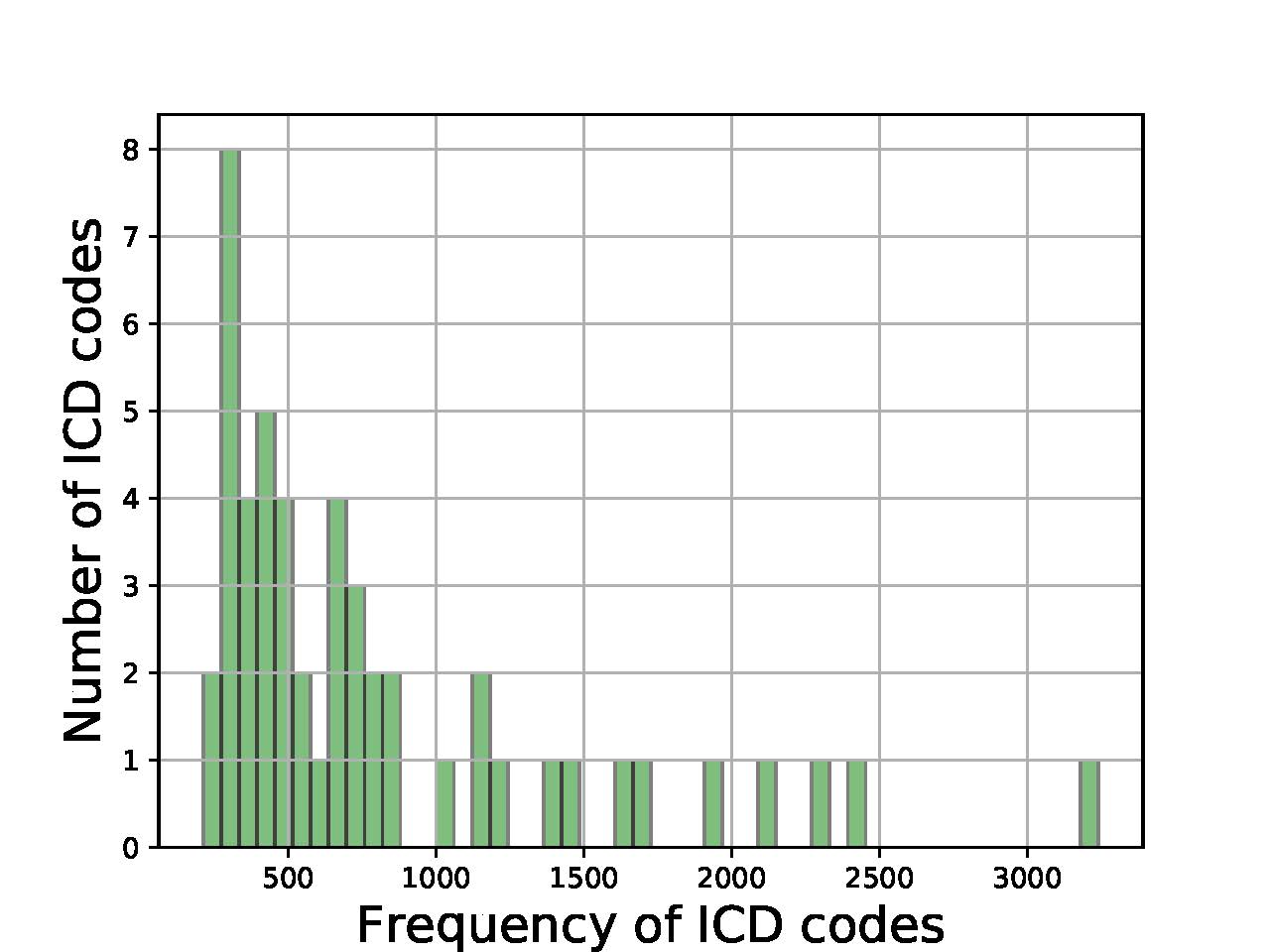}}
  \subfigure[Distribution of ICD code count]{
  	\label{fig:dist:exp:c}
    \includegraphics[width=0.3\linewidth]
    {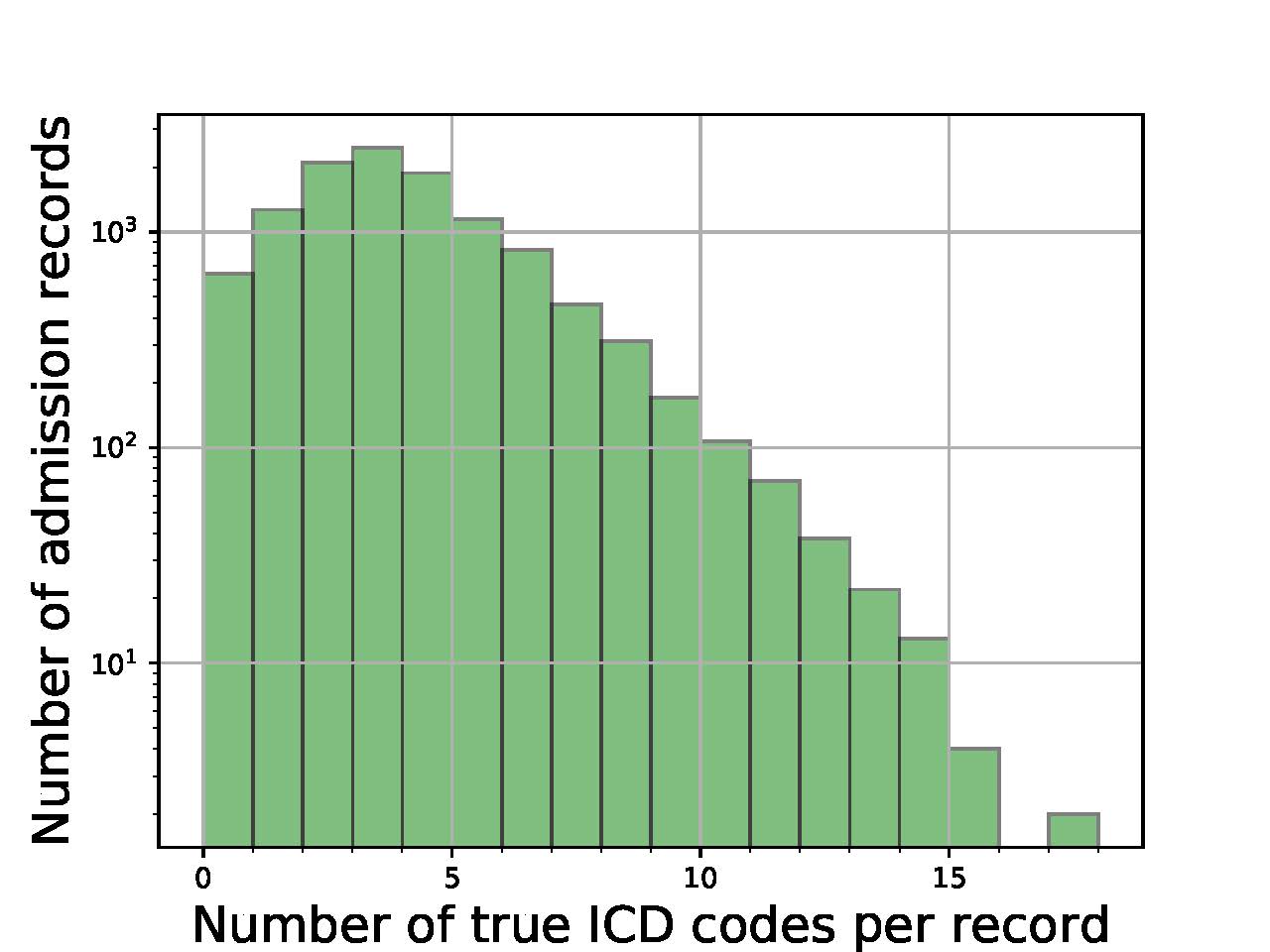}	}
  \caption{Distribution of experimental data}
  \label{fig:dist:exp}
\end{figure}

In this way, we obtain 11,523 hospital admission records with overall 59,302 diagnosis descriptions. Figure \ref{fig:dist:exp:a} shows the distribution of the number of extracted plain-text diagnosis descriptions across medical records. After restricting our ICD coding target to the 50 most frequent codes, the distribution of ICD code frequency is shown in Figure \ref{fig:dist:exp:b}, and the distribution of the number of assigned codes per admission record is shown in \ref{fig:dist:exp:c}. We split the dataset into training set with 8,066 hospital admission records, validation set with 1,728 records, and test set with 1,729 records.

\subsection*{Model design}
\begin{figure}[!ht]
\centering
\includegraphics[width=\linewidth]{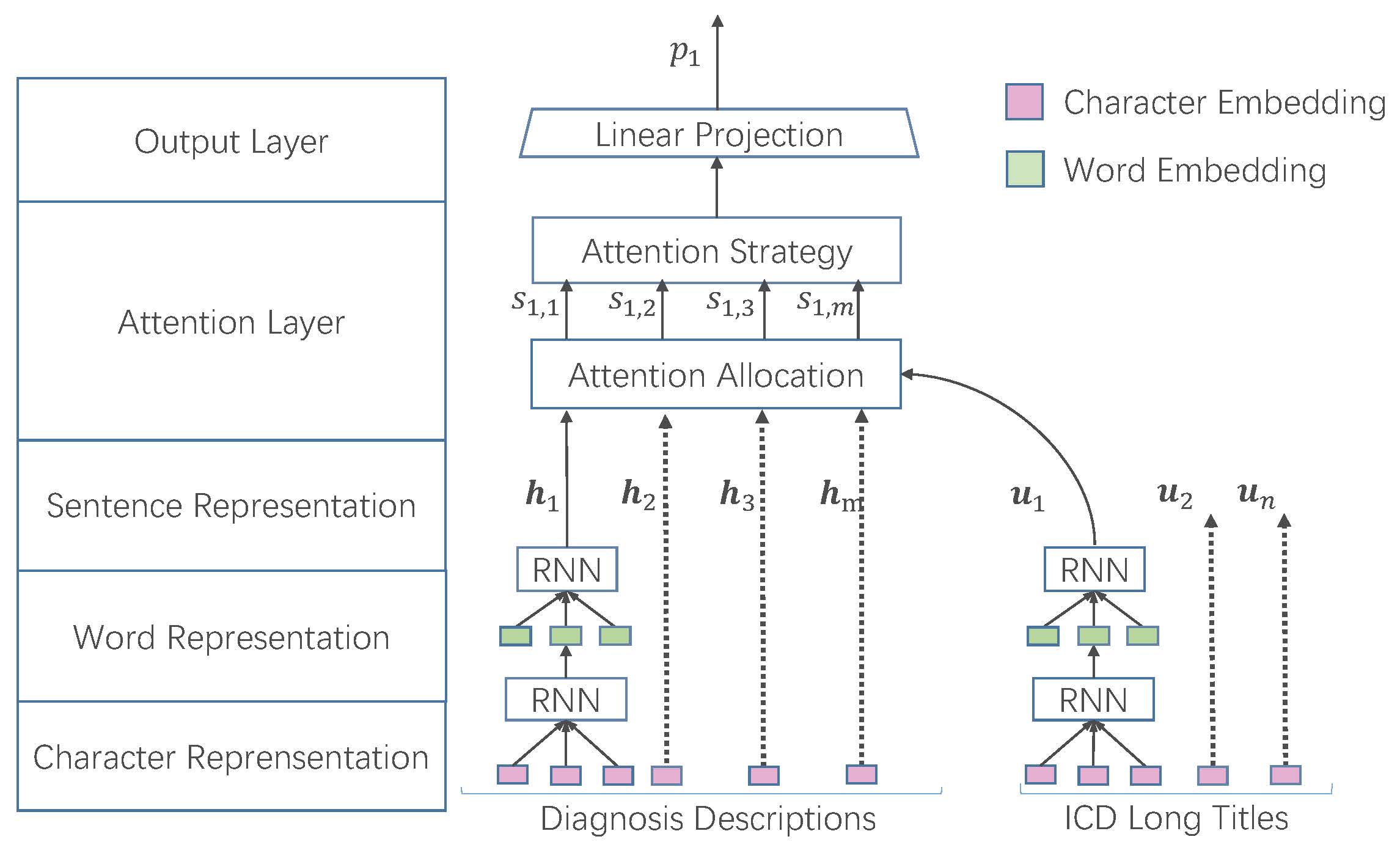}
\caption{Model Architecture.}
\label{fig:modelarch}
\end{figure}

The ICD coding model mainly consists of four modules, which are used for (1) encoding the diagnosis descriptions, (2) encoding the ICD codes based on their textual descriptions, (3) matching diagnosis descriptions with ICD codes, and (4) assigning the ICD codes, respectively. The overall architecture is illustrated in Figure~\ref{fig:modelarch}. In the following we present each component in detail.

\subsubsection*{Diagnosis description encoder}
We leverage the long short-term memory (LSTM) recurrent network to encode the diagnosis descriptions\cite{sundermeyer2012lstm}. LSTM is a popular variant of the recurrent neural network (RNN). Due to the capacity of capturing long-range semantics in texts, LSTM is widely used for language modeling and sequence encoding\cite{mikolov2010recurrent}$^{,}$\cite{bahdanau2014neural}. An LSTM recurrent network consists of a sequence of units, each of which models one item in the input sequence. Each unit consists of an input gate $\bm{i}$, a forget gate $\bm{f}$, a cell gate $\bm{g}$, an output gate $\bm{o}$, a cell state $\bm{c}$, and a hidden state $\bm{h}$, which are all vectors.  They are computed as follows:
\begin{equation}
\begin{array}{ll}
  \bm{i_{t}} = \mathrm{sigmoid}(\bm{W_{ii}} \bm{x_{t}} + \bm{b_{ii}} + \bm{W_{hi}} \bm{h_{t-1}} + \bm{b_{hi}}) \\
  \bm{f_{t}} = \mathrm{sigmoid}(\bm{W_{if}} \bm{x_{t}}+ \bm{b_{if}}+ \bm{W_{hf}} \bm{h_{t-1}} + \bm{b_{hf}}) \\
  \bm{g_{t}} = \tanh(\bm{W}_{ig} \bm{x_{t}} + \bm{b_{ig}} + \bm{W_{hc}} \bm{h_{t-1}} + \bm{b_{hg}}) \\
  \bm{o_{t}} = \mathrm{sigmoid}(\bm{W}_{io} \bm{x_{t}} + \bm{b_{io}} + \bm{W_{ho}} \bm{h_{t-1}} + \bm{b_{ho}}) \\
  \bm{c_{t}} = \bm{f_{t}} * \bm{c_{t-1}} + \bm{i_{t}} * \bm{g_{t}}\\
  \bm{h_{t}} = \bm{o_{t}} * \tanh(\bm{c_{t}})
  \end{array}
  \label{eqt:lstm}
\end{equation}

For clarity, we denote scalars in plain lowercase letters, vectors in bold lowercase, and matrices in bold uppercase. The operator `$*$' in Equation~\ref{eqt:lstm} denotes element-wise multiplication and $t$ represents the time step in the sequence. The sigmoid function is defined as: $\mathrm{sigmoid(x)}=1/(1+exp(-x))$, and the tanh function is $\mathrm{tanh(x)} = (exp(x) - exp(-x)) / (exp(x) + exp(-x))$.

For each diagnosis description, we use both character-level LSTM network and word-level LSTM network to obtain its hidden representation. Specifically, in the character-level LSTM, $\bm{x_{t}}$ is the embedding vector of the $t^{th}$ character in the word, and $T$ is the total number of characters in this word. We select the hidden state of LSTM in the last time step as the hidden representation of the word. In the word-level LSTM, $\bm{x_{t}}$ is the hidden vector of the $t^{th}$ word in the sentence, and $T$ is the number of words. Similarly, we choose the last hidden state as the representation of the sentence. The reason why we choose character-aware encoding method is there are considerable medical terms with same suffix denoting similar diseases and we expect the character-level LSTM to capture such characteristics. In the following, we denote the hidden representations of the written diagnosis descriptions as $\bm{h_{1},h_{2},...,h_{m}}$, where $m$ is the number of extracted diagnosis descriptions in one record. 

\subsubsection*{ICD code encoder}
For each ICD code, we adopt the same two-level LSTM architecture, i.e., character-level and word-level, to obtain the hidden representation of its long title definition, which is provided in the MIMIC-III dataset. For example, in MIMIC-III, the long title of ICD code `4010' is `Malignant essential hypertension'. The hidden vector of `Malignant essential hypertension' obtained with the LSTM network serves as the representation of ICD code `4010'. The parameters of the neural networks for the ICD code encoder and the diagnosis description encoder are not tied, in order to learn different language styles of these two sets of texts. we use $\bm{u_{1},u_{2},...u_{n}}$ to denote the hidden representations of different ICD codes obtained by their long title definitions, where $n$ is the total number of ICD categories. As in our experiment we have picked out the most frequent 50 codes, $n=50$.

\subsubsection*{Attentional match}
Typically, the number of written diagnosis descriptions does not equal to the number of assigned ICD codes, so we cannot directly assign one code to one diagnosis description. Considering that human coders are supposed to assign appropriate codes according to overall health condition, in parallel, we take all diagnosis descriptions into account during coding by adopting an attention strategy. The attention mechanism provides a recipe for choosing which diagnosis descriptions are important when performing coding.

We use $u_{i,k}$ and $h_{j,k}$ to represent the $k^{th}$ dimension of hidden representations of the $i^{th}$ ICD code and the $j^{th}$ diagnosis description, respectively. For the $i^{th}$ ICD code, we use $a_{i,j}$ to denote its attention score on the $j^{th}$ diagnosis description, which is the cosine similarity of the hidden representations of the $i^{th}$ ICD code and the $j^{th}$ diagnosis description.
\begin{equation}
a_{i,j} = \sum_{k=1}^{d}u_{i,k}h_{j,k}
\end{equation}

\begin{figure}[!ht]
\centering
\includegraphics[width=\linewidth]{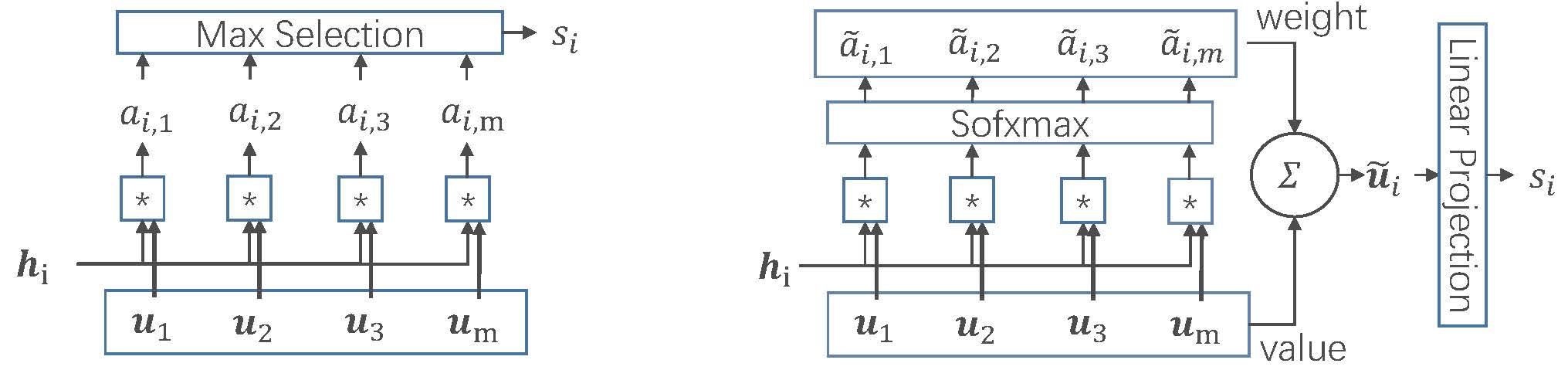}
\caption{Two Architectures of Attentional layer in our model. The `$*$' in the square denotes inner-product calculation. \textbf{Left} is the Hard-selection Mechanism. \textbf{Right} is the Soft-attention Mechanism. }
\label{fig:attention_trick}
\end{figure}

Then we design two different kinds of attention layers to obtain the confidence score of ICD code assignment: {\it Hard-selection} and {\it Soft-attention} mechanism, which are depicted in Figure \ref{fig:attention_trick}.\\

Hard-selection. Based on the assumption that the most related diagnosis description plays a decisive role when assigning ICD code, for each ICD code, we define the dominating diagnosis as the one that has the maximum attention score among all diagnosis descriptions. We apply the sigmoid function to normalize the score into a probability value in $[0, 1]$. The probability of the $i^{th}$ ICD code being assigned is thus:
\begin{equation}
p_{i}=\mathrm{sigmoid}(\max_{j=1,2,...m}(a_{i, j}))
\end{equation}
Soft-attention. Instead of choosing the single maximum attention score, here we apply a softmax function to normalize the attention scores among all diagnosis descriptions into a probability simplex. The normalized attention scores are utilized as the weights of different diagnosis descriptions. We then use the weighted average over the hidden representations of different diagnosis descriptions as the attentional hidden vector. In this way, the attentional hidden vector can take into account all diagnosis descriptions with varying levels of attention. The attentional vector of the $i^{th}$ ICD code is denoted as $\bm{\tilde{u}_{i}}$. 
\begin{equation}
\tilde{a}_{i,j} = \frac{exp(a_{i,j})}{\sum_{j=1}^{m}(exp(a_{i,j}))}
\end{equation}
\begin{equation}
\bm{\tilde{u}}_{i} = \sum_{j=1}^{m}\tilde{a}_{i,j}*\bm{h_{j}}
\end{equation}

\subsubsection*{Linear projection layer}
For the attentional hidden vector $\bm{\tilde{u}}_{i}$, we use linear Perceptron structure as the output layer to project the vector into a real value\cite{rosenblatt1958perceptron}$^{,}$\cite{widrow199030}, which represents the confidence score of predicting label to be true. The Perceptron parameters are different among each code. Finally, we utilize sigmoid function to normalize the confidence score into a probability, which ranges from 0 to 1.
\begin{equation}
s_{i} = \sum_{k=1}^{d}w_{i,k}\tilde{u}_{i,k}
\end{equation}
\begin{equation}
p_{i} = \mathrm{sigmoid}(s_{i})
\end{equation}

\subsubsection*{Parameter learning}
We use binary cross entropy as the loss function for each ICD code\cite{deng2006cross,cover2012elements}. The loss function for each pat record can be formulated as follows:
\begin{equation}
Loss = -1/n\sum_{j=1}^{n}(t_{i}*log(p_{i})+(1-t_{i})*log(1-p_{i})),
\end{equation}
where $t_i$ is the real label of the $i^{th}$ ICD code, i.e., true(1) or false(0). All parameters are learned by minimizing the loss function with stochastic gradient descent\cite{bottou2010large}.

\subsubsection*{Hyperparameter setting}
The model is trained on the training set using the standard ADAM optimizer\cite{kingma2014adam}, with an initial learning rate 0.001 and mini-batch size 10. Hyper-parameters are fine-tuned on the validation set. In particular, the number of hidden units and output units of all LSTM modules are 200. For word-level LSTM in our experiment, we apply a dropout layer with 0.5 dropout probability to the output, to avoid the overfitting problem\cite{srivastava2014dropout}. Since our model provides a probability score for each assignment of ICD code, we also tune on the validation set the optimal threshold that cuts the probability score into a binary output, i.e., true or false, to obtain best F1 score.

\subsection*{Analysis and evaluation}
Considering the ICD code assignment is generally sparse, with most ICD codes labeled as false and only a few as true, we use the micro F1-score and AUC\_ROC (area under curve of receiver operating characteristic) score as the quantitative metrics. Micro F1-score is a harmonic mean of precision and recall. It is widely used to evaluate the performance of a binary classifier on imbalanced data\cite{van1979information}. The micro AUC\_ROC score is calculated as the area under the ROC curve, which is drawn by plotting the true positive rate (TPR) against the false positive rate (FPR) at various threshold settings\cite{hanley1982meaning}$^{,}$\cite{fawcett2006introduction}. Intuitively, the AUC\_ROC score measures the probability that the model assigns higher score for a positive instance than negative one, the lower bound of which is $0.5$.

\section*{Results}
Table \ref{tab:performance} shows the F1-score and AUC\_ROC score of different models evaluated on the test set. We can observe that with the Soft-attention mechanism, the F1 and AUC\_ROC increase 5.2 and 2.3 percent, respectively, compared to the Hard-selection model. To further explore the efficacy of different modules in our model, we perform an ablation study on our intact Soft-attention model, which leverages character-level and word-level LSTM encoder and Soft-attention mechanism. Performance decreasing on several ablation experiments indicates that all the designed modules in our model are necessary and play a crucial role in the coding process. 
\begin{table}[!h]
  \centering
\begin{tabular}{|l|l|r|r|}
\hline
\multicolumn{2}{|c|}{Model Architecture} & \multicolumn{1}{l|}{F1} & \multicolumn{1}{l|}{AUC\_ROC} \bigstrut\\
\hline
\multicolumn{2}{|l|}{Hard-selection Model} & 0.480& 0.877\bigstrut\\
\hline
\multicolumn{2}{|l|}{Soft-attention Model} & \textbf{0.532} & \textbf{0.900} \bigstrut\\
\hline
\multicolumn{4}{c}{} \bigstrut\\
\hline
\multicolumn{4}{|c|}{Ablation Study on Soft-attention Model} \bigstrut\\
\hline
\multicolumn{2}{|l|}{Replace character-level LSTM with random initialized and tunable word embedding} & 0.508 & 0.882 \bigstrut\\
\hline
\multicolumn{2}{|l|}{Replace character-level LSTM with pre-trained and tunable word embedding} & 0.528 & 0.895 \bigstrut\\
\hline
\multicolumn{2}{|l|}{Replace word-level LSTM encoder with average encoder} & 0.504 & 0.886 \bigstrut\\
\hline
\multicolumn{2}{|l|}{Replace attention mechanism with naïve linear classifier} & 0.471 & 0.882 \bigstrut\\
\hline
\end{tabular}%
   \caption{Performance on different models}
  \label{tab:performance}%
\end{table}%
The attention scores on the subset of 50 ICD codes is shown in Table~\ref{tab:attention} for one patient visit sample. We can observe that for different ICD codes, our model allocates different attention scores to diagnosis descriptions automatically. For example, when assigning the ICD code titled `Neonatal jaundice associated with preterm delivery', the model puts more attention on `Prematurity at 34 and 5/7 weeks gestation' and `Hyperbilirubinemia of prematurity', while less attention on other irrelevant diagnosis descriptions like `Sepsis ruled out'. The attention allocation results of all 50 the ICD codes for this sample can be found in the supplementary materials.
\begin{table}[!htbp]
  \centering
\begin{tabular}{|l|l|l|l|l|l|}
\hline
Index & \multicolumn{5}{l|}{Diagnosis description} \bigstrut\\
\hline
1     & \multicolumn{5}{l|}{Prematurity at 34 and 5/7 weeks gestation} \bigstrut\\
\hline
2    & \multicolumn{5}{l|}{Twin \# 2} \bigstrut\\
\hline
3     & \multicolumn{5}{l|}{Status post transitional respiratory distress} \bigstrut\\
\hline
4     & \multicolumn{5}{l|}{Sepsis ruled out} \bigstrut\\
\hline
5     & \multicolumn{5}{l|}{Hyperbilirubinemia of prematurity} \bigstrut\\
\hline
\end{tabular}%

\begin{tabular}{|p{33.065em}|r|r|r|r|r|}
\multicolumn{1}{r}{} & \multicolumn{1}{r}{} & \multicolumn{1}{r}{} & \multicolumn{1}{r}{} & \multicolumn{1}{r}{} & \multicolumn{1}{r}{} \bigstrut[b] \\
\hline
LONG TITLE OF ICD CODE & 1     & 2     & 3     & 4     & 5 \bigstrut\\
\hline
Esophageal reflux & 0.20  & 0.09  & 0.12  & \textbf{0.52 } & 0.07  \bigstrut\\
\hline
Acute kidney failure with lesion of tubular necrosis & \textbf{0.48 } & 0.14  & 0.03  & 0.08  & 0.27  \bigstrut\\
\hline
Acute kidney failure, unspecified & 0.43  & 0.11  & 0.04  & 0.11  & 0.31  \bigstrut\\
\hline
Chronic kidney disease, unspecified & \textbf{0.44 } & 0.21  & 0.02  & 0.18  & 0.16  \bigstrut\\
\hline
Urinary tract infection, site not specified & 0.24  & 0.06  & 0.01  & \textbf{0.65 } & 0.04  \bigstrut\\
\hline
\textbf{Neonatal jaundice associated with preterm delivery} & \textbf{0.57 } & 0.14  & 0.00  & 0.00  & 0.28  \bigstrut\\
\hline
Septic shock & \textbf{0.58 } & 0.21  & 0.00  & 0.12  & 0.09  \bigstrut\\
\hline
Severe sepsis & \textbf{0.57 } & 0.16  & 0.00  & 0.16  & 0.09  \bigstrut\\
\hline
Cardiac complications, not elsewhere classified & 0.35  & \textbf{0.41 } & 0.00  & 0.13  & 0.11  \bigstrut\\
\hline
Aortocoronary bypass status & 0.37  & \textbf{0.38 } & 0.02  & 0.10  & 0.14  \bigstrut\\
\hline
Percutaneous transluminal coronary angioplasty status & 0.31  & \textbf{0.45 } & 0.02  & 0.07  & 0.14  \bigstrut\\
\hline
Long-term (current) use of anticoagulants & 0.23  & \textbf{0.51 } & 0.02  & 0.11  & 0.13  \bigstrut\\
\hline
Long-term (current) use of insulin & 0.24  & \textbf{0.31 } & 0.05  & 0.25  & 0.15  \bigstrut\\
\hline
\textbf{Observation for suspected infectious condition} & 0.16  & 0.10  & 0.10  & \textbf{0.56 } & 0.08  \bigstrut\\
\hline
Single liveborn, born in hospital, delivered without mention of cesarean section & 0.02  & \textbf{0.96 } & 0.00  & 0.01  & 0.01  \bigstrut\\
\hline
Single liveborn, born in hospital, delivered by cesarean section & 0.02  & \textbf{0.95 } & 0.00  & 0.01  & 0.02  \bigstrut\\
\hline
\textbf{Need for prophylactic vaccination and inoculation against viral hepatitis} & 0.34  & \textbf{0.38 } & 0.02  & 0.10  & 0.16  \bigstrut\\
\hline
Personal history of tobacco use & \textbf{0.52 } & 0.19  & 0.01  & 0.17  & 0.11  \bigstrut\\
\hline

\end{tabular}%
  \caption{Attention allocation on one sample. \textbf{Top} shows all the extracted diagnosis descriptions from written discharge diagnosis given by physicians. \textbf{Down} shows the attention allocation on a subset of ICD codes. The codes in bold format have true labels.}
  \label{tab:attention}%
\end{table}%
\section*{Discussion}

Our model achieves attractive performance on the ICD coding task, which indicates our architecture is reasonable: to extract diagnosis descriptions from discharge summary, and use the hidden meaning of these descriptions to predict on target ICD code whose meaning is represented by its formal title, can be a promising methodology to perform automated ICD coding. The Soft-attention mechanism can push the model to allocate different attention on multiple diagnosis descriptions, which can obtain better performance compared to Hard-selection. In the following, we will discuss several important insights provided by our ablation study on the Soft-attention model in detail.

To evaluate the effectiveness of the character-level LSTM module in learning the hidden representation of medical vocabulary, we remove it from the model. Instead, we obtain the hidden vector of each word from a tunable word embedding layer, where each word is assigned a fixed-dimension vector. Note that the other modules remain intact and the Soft-attention strategy is leveraged. It causes 0.024 and 0.018 drop of the F1 score and AUC\_ROC respectively, which demonstrates the necessity of incorporating character-level encoder into representation learning. We have also tried to initialize the word embedding layer with pre-trained word vectors. These vectors were learned using word2vec tool on a large corpus of medical research papers collected from Pubmed, BioMed and PLOS\cite{bengio2003neural}. In this setting, All the words are transformed to lowercase and lemmatized in advance. The performance has increased compared to randomly initialized word embeddings but it's still lower than our character-level LSTM model. 

To ensure our character-level LSTM module can give reasonable representations for diagnosis description, we have checked the nearest neighbors of words and sentences based on Euclidean distance in hidden space. Table \ref{tab:hiddenlocation} shows a subset of words and sentences and their nearest neighbors. On top of the table displays the word neighbor relationship contrast between the model with character-level LSTM and word embedding layer with pre-trained word vectors. First, it indicates that character-level LSTM word encoder can correct various typos and recognize different morphologies appearing in the written diagnosis descriptions, by generating similar representations for them. For example, our model can recognize different written variants of `Ischemia' and generate near representations for them. In addition, many disease names and procedures with same suffix are denoting similar diseases, which can be captured by our character-level LSTM encoder efficiently. Otherwise, there exist some words with same suffix but unrelated meanings indeed that are also distributed near in the hidden space, however, these words are not denoting disease categories in many cases, like `state', so it should have little effect to the coding. While looking at the sentence neighbor relationship shown at the bottom of the table \ref{tab:hiddenlocation}, we could observe that out model can generate near embeddings for similar sentences too.
\begin{table}[!htbp]
  \centering
\begin{tabular}{|p{8.5em}|c|p{30em}|}
\hline
\multicolumn{2}{|p{14.5em}|}{Word encoding method} & Neighbors \bigstrut\\
\hline
Character-level LSTM & \multicolumn{1}{p{6em}|}{Ischemia} & \textit{Ischmia} (2.76), ischemia (4.05), Dysthymia (5.46), hypercalcemia (5.74), \textit{ishemia} (5.76) \bigstrut\\
\cline{2-3}\multicolumn{1}{|c|}{} & \multicolumn{1}{p{6em}|}{Pneumonia} & \textit{Penumonia} (0.37), \textit{Pnuemonia} (0.85), \textit{pnuemonia} (1.77), \textit{Pnemonia} (1.78), pneumonia (1.95) \bigstrut\\
\cline{2-3}\multicolumn{1}{|c|}{} & \multicolumn{1}{p{6em}|}{Gastroenteritis} & gastroenteritis (2.59), Osteoarthritis (3.99), endomyometritis (4.18), Gastritis (4.18), prostatitis (4.28) \bigstrut\\
\cline{2-3}\multicolumn{1}{|c|}{} & \multicolumn{1}{p{6em}|}{Coronary} & coronary (2.04), Rotary (4.47), ovary (6.60), Artery (7.16), aortopulmonary (7.16) \bigstrut\\
\cline{2-3}\multicolumn{1}{|c|}{} & \multicolumn{1}{p{6em}|}{State} & state (4.04), resuscitate (4.73), prostate (4.81), lactate (4.99), Methotrexate (5.06) \bigstrut\\
\hline
Word2vec & \multicolumn{1}{p{6em}|}{ischemia} & ischemic (12.51), reperfusion (15.68), malperfusion (17.87), infarction (18.10), Microinfarction (18.39) \bigstrut\\
\cline{2-3}\multicolumn{1}{|c|}{} & \multicolumn{1}{p{6em}|}{pneumonia} & tracheobronchitis (16.02), tracheitis (16.32), epiglottitis (16.84), abcess (17.07), septicemia (17.09) \bigstrut\\
\cline{2-3}\multicolumn{1}{|c|}{} & \multicolumn{1}{p{6em}|}{gastroenteritis} & diarrhea (17.43), tracheitis (17.90), enteritis (17.91), stools (17.98), diarrheal (18.10) \bigstrut\\
\cline{2-3}\multicolumn{1}{|c|}{} & \multicolumn{1}{p{6em}|}{coronary} & multivessel (16.80), vessels (16.87), atherectomy (16.96), aortoiliac (17.15), pectoris (17.32) \bigstrut\\
\cline{2-3}\multicolumn{1}{|c|}{} & \multicolumn{1}{p{6em}|}{state} & the (15.44), and (15.66), a (15.67), transition (15.71), pauses (15.92), of/Of (16.12) \bigstrut\\
\cline{2-3}\multicolumn{1}{|c|}{} & \multicolumn{1}{p{6em}|}{\textit{ishemia}} & 4.3 (8.33), t4-t5 (8.39), d'or (8.40), bronchiectesis (8.44), difficule (8.44) \bigstrut\\
\cline{2-3}\multicolumn{1}{|c|}{} & \multicolumn{1}{p{6em}|}{\textit{pnuemonia}} & 10\% (8.28), 76 (8.35), enterovaginal (8.36), penicillion (8.39), secudnum (8.40) \bigstrut\\
\hline
\multicolumn{3}{c}{} \bigstrut\\
\end{tabular}
\begin{tabular}{|p{14.5em}|p{31.875em}|}
\hline
Diagnosis descriptions & Neighbors \bigstrut\\
\hline
Coronary artery disease & Coronary Artery Disease (0.78), Acute coronary artery disease (1.36), Coronary artery disease stable (1.75), Chronic coronary artery disease (1.82), History of coronary artery disease (1.88) \bigstrut\\
\hline
Congestive heart failure & congestive heart failure (1.66), Pulmonary hypertension / congestive heart failure (1.76), Diastolic congestive heart failure (1.96), Biventricular congestive heart failure (1.99), Diastolic Dysfunction Heart Failure (2.04) \bigstrut\\
\hline
Hemodynamic monitoring with central venous catheter & Bladder atony status post suprapubic catheter (2.33), Patient has suprapubic catheter (2.48), hepatic dysfunction which is resolving (2.70), Erosions in the stomach and duodenum (2.72), ? Gastrointestinal bleed (2.76) \bigstrut\\
\hline
Apnea of prematurity & Apnea of Prematurity (0.27), Apnea of prematurity ongoing (1.41), Retinopathy of prematurity (1.45), Apnea and bradycardia of prematurity (1.47), Apnea bradycardia of prematurity (1.56) \bigstrut\\
\hline
Diabetes mellitus type 2 & Diabetes mellitus , type 2 (0.35), Diabetes mellitus Type 2 (0.52), Diabetes mellitus , Type 2 (0.73), Diabetes mellitus 2 (1.17), Diabetes Type 2 (1.20) \bigstrut\\
\hline
\end{tabular}
    \caption{Hidden vector location relationship. \textbf{Top} is a comparison between the character-level LSTM encoder and the word2vec model with pre-trained word vectors. The italic text means typo appearing in the diagnosis descriptions. \textbf{Bottom} is the diagnosis neighbor relation in our intact Soft-attention model.}
  \label{tab:hiddenlocation}
\end{table}

Besides word-level LSTM encoder, word averaging method also shows strong performance to generate sentence embeddings\cite{wieting2015towards}. we have also tried averaging the word embeddings in one sentence to obtain the hidden vector of sentence, instead of using LSTM encoder. Keeping other modules intact, the F1 drops 0.028 and AUC\_ROC drops 0.014, which indicates the word-level LSTM encoder is superior to word averaging.

We evaluate the necessity of the attention mechanism by comparing our Soft-attention model with a linear classifier without attention mechanism. We design the architecture of linear classifier as follows. For each ICD code, we concatenate its hidden vector with the representation of the overall diagnosis, which is obtained by averaging the hidden vectors of each diagnosis description. The concatenated vector is processed by a linear Perceptron to get the confidence score. The parameters of linear Perceptron are independent among different ICD codes. Replacing attention mechanism with such a linear classifier causes the F1 and AUC\_ROC to drop 0.061 and 0.018 respectively, which demonstrates the advantage of our attention mechanism.

\section*{Limitations}
The performance achieved by our hierarchical neural models with Soft-attention mechanism shows that reliable performance could be obtained even through a simple diagnosis description extraction process. But, considering the noisy format of the electrical discharge summary, with a more elaborate diagnosis extraction preprocessing and cleaner corpus with high-quality diagnosis descriptions, we believe the performance could be improved further. 

Another limitation of our study is the candidate ICD codes are restricted to the most frequent 50 ones. If one ICD code is too rare, there will not be enough evidence to construct a valid neural model, and the label imbalance problem will be more severe\cite{japkowicz2002class}, which makes the learning harder. It should be helpful to obtain more formatted records to support the model to learn.

With separate linear Perceptrons to assign each ICD code, we have assumed that the assignment of different ICD codes are mutually independent. However, ICD codes indeed correlate with each other to some extent. For example, the long title name of ICD code `40390' is `Hypertensive chronic kidney disease, unspecified, with chronic kidney disease stage I through stage IV, or unspecified', while `40391' is `Hypertensive chronic kidney disease, unspecified, with chronic kidney disease stage V or end stage renal disease'. If ICD code `40390' is assigned to one patient record, ICD `40391' should definitely not be assigned since these two codes represent exclusive health conditions. And it might be helpful to leverage the hierarchy structure of ICD codes\cite{perotte2013diagnosis}. Thus, modeling such correlations with some structured methods can be meaningful for improving performance.

\section*{Conclusions}
We find it is promising to construct a high quality ICD coding machine directly from diagnosis description in the electronic medical records. Our model achieves high performance, suggesting that an attentional match between the diagnosis descriptions and the textual definition of ICD code suits well for the inference task. The proposed Soft-attention mechanism can learn to allocate varying attention strengths on multiple diagnosis descriptions when assigning ICD codes. Just like the reasoning of human, with more attention on informative diagnosis descriptions and less on irrelevant ones, the Soft-attention model can assign codes based on diagnosis descriptions automatically and efficiently. 

Our experiment indicates the potential for real life applications in view of the high performance even on some noisy-formatted data. We believe that with more elaborate data preprocessing techniques, and with more formatted electrical medical records, the automatic coding can be even more accurate. Since our model can give a probability score when assigning ICD code, we can decrease the probability threshold to get higher recall rate or increase to get higher precision. Thus, in addition to coding ICD diagnosis directly, our model can also serve as an assistant tool for doctors, helping them to pre-select a small set of candidate codes and thus greatly alleviating doctors' workloads.

In this paper we have adopted ICD-9 diagnostic codes as coding target, however the proposed approach can straightforwardly be adapted to new revisions of ICD codes, like ICD-10\cite{world1992international}, as long as the formal definitions of all codes and golden diagnostic codes on training data are available.

\bibliography{reference}

\begin{thebibliography}{10}
\expandafter\ifx\csname url\endcsname\relax
  \def\url#1{\texttt{#1}}\fi
\expandafter\ifx\csname urlprefix\endcsname\relax\def\urlprefix{URL }\fi
\expandafter\ifx\csname doiprefix\endcsname\relax\def\doiprefix{DOI }\fi
\providecommand{\bibinfo}[2]{#2}
\providecommand{\eprint}[2][]{\url{#2}}

\bibitem{world1978international}
\bibinfo{author}{Organization, W.~H.} \emph{et~al.}
\newblock \bibinfo{journal}{\bibinfo{title}{International classification of
  diseases:[9th] ninth revision, basic tabulation list with alphabetic index}}.
\newblock {\emph{\JournalTitle{World Health Organization}}}
  (\bibinfo{year}{1978}).

\bibitem{o2005measuring}
\bibinfo{author}{O'malley, K.~J.} \emph{et~al.}
\newblock \bibinfo{journal}{\bibinfo{title}{Measuring diagnoses: Icd code
  accuracy}}.
\newblock {\emph{\JournalTitle{Health services research}}}
  \textbf{\bibinfo{volume}{40}}, \bibinfo{pages}{1620--1639}
  (\bibinfo{year}{2005}).

\bibitem{sheppard2008ambiguous}
\bibinfo{author}{Sheppard, J.~E.}, \bibinfo{author}{Weidner, L.~C.},
  \bibinfo{author}{Zakai, S.}, \bibinfo{author}{Fountain-Polley, S.} \&
  \bibinfo{author}{Williams, J.}
\newblock \bibinfo{journal}{\bibinfo{title}{Ambiguous abbreviations: an audit
  of abbreviations in paediatric note keeping}}.
\newblock {\emph{\JournalTitle{Archives of disease in childhood}}}
  \textbf{\bibinfo{volume}{93}}, \bibinfo{pages}{204--206}
  (\bibinfo{year}{2008}).

\bibitem{lang2007consultant}
\bibinfo{author}{Lang, D.}
\newblock \bibinfo{journal}{\bibinfo{title}{Consultant report-natural language
  processing in the health care industry}}.
\newblock {\emph{\JournalTitle{Cincinnati Children's Hospital Medical Center,
  Winter}}}  (\bibinfo{year}{2007}).

\bibitem{farkas2008automatic}
\bibinfo{author}{Farkas, R.} \& \bibinfo{author}{Szarvas, G.}
\newblock \bibinfo{journal}{\bibinfo{title}{Automatic construction of
  rule-based icd-9-cm coding systems}}.
\newblock {\emph{\JournalTitle{BMC bioinformatics}}}
  \textbf{\bibinfo{volume}{9}}, \bibinfo{pages}{S10} (\bibinfo{year}{2008}).

\bibitem{spyns1996natural}
\bibinfo{author}{Spyns, P.}
\newblock \bibinfo{journal}{\bibinfo{title}{Natural language processing}}.
\newblock {\emph{\JournalTitle{Methods of information in medicine}}}
  \textbf{\bibinfo{volume}{35}}, \bibinfo{pages}{285--301}
  (\bibinfo{year}{1996}).

\bibitem{hearst1999untangling}
\bibinfo{author}{Hearst, M.~A.}
\newblock \bibinfo{title}{Untangling text data mining}.
\newblock In \emph{\bibinfo{booktitle}{Proceedings of the 37th annual meeting
  of the Association for Computational Linguistics on Computational
  Linguistics}}, \bibinfo{pages}{3--10} (\bibinfo{organization}{Association for
  Computational Linguistics}, \bibinfo{year}{1999}).

\bibitem{zeng2006extracting}
\bibinfo{author}{Zeng, Q.~T.} \emph{et~al.}
\newblock \bibinfo{journal}{\bibinfo{title}{Extracting principal diagnosis,
  co-morbidity and smoking status for asthma research: evaluation of a natural
  language processing system}}.
\newblock {\emph{\JournalTitle{BMC medical informatics and decision making}}}
  \textbf{\bibinfo{volume}{6}}, \bibinfo{pages}{30} (\bibinfo{year}{2006}).

\bibitem{ananiadou2006text}
\bibinfo{author}{Ananiadou, S.} \& \bibinfo{author}{McNaught, J.}
\newblock \emph{\bibinfo{title}{Text mining for biology and biomedicine}}
  (\bibinfo{publisher}{Artech House London}, \bibinfo{year}{2006}).

\bibitem{pestian2007shared}
\bibinfo{author}{Pestian, J.~P.} \emph{et~al.}
\newblock \bibinfo{title}{A shared task involving multi-label classification of
  clinical free text}.
\newblock In \emph{\bibinfo{booktitle}{Proceedings of the Workshop on BioNLP
  2007: Biological, Translational, and Clinical Language Processing}},
  \bibinfo{pages}{97--104} (\bibinfo{organization}{Association for
  Computational Linguistics}, \bibinfo{year}{2007}).

\bibitem{meystre2008extracting}
\bibinfo{author}{Meystre, S.~M.}, \bibinfo{author}{Savova, G.~K.},
  \bibinfo{author}{Kipper-Schuler, K.~C.}, \bibinfo{author}{Hurdle, J.~F.}
  \emph{et~al.}
\newblock \bibinfo{journal}{\bibinfo{title}{Extracting information from textual
  documents in the electronic health record: a review of recent research}}.
\newblock {\emph{\JournalTitle{Yearb Med Inform}}}
  \textbf{\bibinfo{volume}{35}}, \bibinfo{pages}{44} (\bibinfo{year}{2008}).

\bibitem{koopman2015automatic}
\bibinfo{author}{Koopman, B.}, \bibinfo{author}{Zuccon, G.},
  \bibinfo{author}{Nguyen, A.}, \bibinfo{author}{Bergheim, A.} \&
  \bibinfo{author}{Grayson, N.}
\newblock \bibinfo{journal}{\bibinfo{title}{Automatic icd-10 classification of
  cancers from free-text death certificates}}.
\newblock {\emph{\JournalTitle{International journal of medical informatics}}}
  \textbf{\bibinfo{volume}{84}}, \bibinfo{pages}{956--965}
  (\bibinfo{year}{2015}).

\bibitem{larkey1996combining}
\bibinfo{author}{Larkey, L.~S.} \& \bibinfo{author}{Croft, W.~B.}
\newblock \bibinfo{title}{Combining classifiers in text categorization}.
\newblock In \emph{\bibinfo{booktitle}{Proceedings of the 19th annual
  international ACM SIGIR conference on Research and development in information
  retrieval}}, \bibinfo{pages}{289--297} (\bibinfo{organization}{ACM},
  \bibinfo{year}{1996}).

\bibitem{franz2000automated}
\bibinfo{author}{Franz, P.}, \bibinfo{author}{Zaiss, A.},
  \bibinfo{author}{Schulz, S.}, \bibinfo{author}{Hahn, U.} \&
  \bibinfo{author}{Klar, R.}
\newblock \bibinfo{title}{Automated coding of diagnoses--three methods
  compared.}
\newblock In \emph{\bibinfo{booktitle}{Proceedings of the AMIA Symposium}},
  \bibinfo{pages}{250} (\bibinfo{organization}{American Medical Informatics
  Association}, \bibinfo{year}{2000}).

\bibitem{perotte2013diagnosis}
\bibinfo{author}{Perotte, A.} \emph{et~al.}
\newblock \bibinfo{journal}{\bibinfo{title}{Diagnosis code assignment: models
  and evaluation metrics}}.
\newblock {\emph{\JournalTitle{Journal of the American Medical Informatics
  Association}}} \textbf{\bibinfo{volume}{21}}, \bibinfo{pages}{231--237}
  (\bibinfo{year}{2013}).

\bibitem{johnson2016mimic}
\bibinfo{author}{Johnson, A.~E.} \emph{et~al.}
\newblock \bibinfo{journal}{\bibinfo{title}{Mimic-iii, a freely accessible
  critical care database}}.
\newblock {\emph{\JournalTitle{Scientific data}}} \textbf{\bibinfo{volume}{3}}
  (\bibinfo{year}{2016}).

\bibitem{prakash2017condensed}
\bibinfo{author}{Prakash, A.} \emph{et~al.}
\newblock \bibinfo{title}{Condensed memory networks for clinical diagnostic
  inferencing.}
\newblock In \emph{\bibinfo{booktitle}{AAAI}}, \bibinfo{pages}{3274--3280}
  (\bibinfo{year}{2017}).

\bibitem{sundermeyer2012lstm}
\bibinfo{author}{Sundermeyer, M.}, \bibinfo{author}{Schl{\"u}ter, R.} \&
  \bibinfo{author}{Ney, H.}
\newblock \bibinfo{title}{Lstm neural networks for language modeling}.
\newblock In \emph{\bibinfo{booktitle}{Thirteenth Annual Conference of the
  International Speech Communication Association}} (\bibinfo{year}{2012}).

\bibitem{mikolov2010recurrent}
\bibinfo{author}{Mikolov, T.}, \bibinfo{author}{Karafi{\'a}t, M.},
  \bibinfo{author}{Burget, L.}, \bibinfo{author}{Cernock{\`y}, J.} \&
  \bibinfo{author}{Khudanpur, S.}
\newblock \bibinfo{title}{Recurrent neural network based language model.}
\newblock In \emph{\bibinfo{booktitle}{Interspeech}}, vol.~\bibinfo{volume}{2},
  \bibinfo{pages}{3} (\bibinfo{year}{2010}).

\bibitem{bahdanau2014neural}
\bibinfo{author}{Bahdanau, D.}, \bibinfo{author}{Cho, K.} \&
  \bibinfo{author}{Bengio, Y.}
\newblock \bibinfo{journal}{\bibinfo{title}{Neural machine translation by
  jointly learning to align and translate}}.
\newblock {\emph{\JournalTitle{arXiv preprint arXiv:1409.0473}}}
  (\bibinfo{year}{2014}).

\bibitem{rosenblatt1958perceptron}
\bibinfo{author}{Rosenblatt, F.}
\newblock \bibinfo{journal}{\bibinfo{title}{The perceptron: A probabilistic
  model for information storage and organization in the brain.}}
\newblock {\emph{\JournalTitle{Psychological review}}}
  \textbf{\bibinfo{volume}{65}}, \bibinfo{pages}{386} (\bibinfo{year}{1958}).

\bibitem{widrow199030}
\bibinfo{author}{Widrow, B.} \& \bibinfo{author}{Lehr, M.~A.}
\newblock \bibinfo{journal}{\bibinfo{title}{30 years of adaptive neural
  networks: perceptron, madaline, and backpropagation}}.
\newblock {\emph{\JournalTitle{Proceedings of the IEEE}}}
  \textbf{\bibinfo{volume}{78}}, \bibinfo{pages}{1415--1442}
  (\bibinfo{year}{1990}).

\bibitem{deng2006cross}
\bibinfo{author}{Deng, L.-Y.}
\newblock \bibinfo{title}{The cross-entropy method: a unified approach to
  combinatorial optimization, monte-carlo simulation, and machine learning}
  (\bibinfo{year}{2006}).

\bibitem{cover2012elements}
\bibinfo{author}{Cover, T.~M.} \& \bibinfo{author}{Thomas, J.~A.}
\newblock \emph{\bibinfo{title}{Elements of information theory}}
  (\bibinfo{publisher}{John Wiley \& Sons}, \bibinfo{year}{2012}).

\bibitem{bottou2010large}
\bibinfo{author}{Bottou, L.}
\newblock \bibinfo{title}{Large-scale machine learning with stochastic gradient
  descent}.
\newblock In \emph{\bibinfo{booktitle}{Proceedings of COMPSTAT'2010}},
  \bibinfo{pages}{177--186} (\bibinfo{publisher}{Springer},
  \bibinfo{year}{2010}).

\bibitem{kingma2014adam}
\bibinfo{author}{Kingma, D.} \& \bibinfo{author}{Ba, J.}
\newblock \bibinfo{journal}{\bibinfo{title}{Adam: A method for stochastic
  optimization}}.
\newblock {\emph{\JournalTitle{arXiv preprint arXiv:1412.6980}}}
  (\bibinfo{year}{2014}).

\bibitem{srivastava2014dropout}
\bibinfo{author}{Srivastava, N.}, \bibinfo{author}{Hinton, G.~E.},
  \bibinfo{author}{Krizhevsky, A.}, \bibinfo{author}{Sutskever, I.} \&
  \bibinfo{author}{Salakhutdinov, R.}
\newblock \bibinfo{journal}{\bibinfo{title}{Dropout: a simple way to prevent
  neural networks from overfitting.}}
\newblock {\emph{\JournalTitle{Journal of machine learning research}}}
  \textbf{\bibinfo{volume}{15}}, \bibinfo{pages}{1929--1958}
  (\bibinfo{year}{2014}).

\bibitem{van1979information}
\bibinfo{author}{Van~Rijsbergen, C.}
\newblock \bibinfo{journal}{\bibinfo{title}{Information retrieval. dept. of
  computer science, university of glasgow}}.
\newblock {\emph{\JournalTitle{URL: citeseer.ist.psu.
  edu/vanrijsbergen79information.html}}} \textbf{\bibinfo{volume}{14}}
  (\bibinfo{year}{1979}).

\bibitem{hanley1982meaning}
\bibinfo{author}{Hanley, J.~A.} \& \bibinfo{author}{McNeil, B.~J.}
\newblock \bibinfo{journal}{\bibinfo{title}{The meaning and use of the area
  under a receiver operating characteristic (roc) curve.}}
\newblock {\emph{\JournalTitle{Radiology}}} \textbf{\bibinfo{volume}{143}},
  \bibinfo{pages}{29--36} (\bibinfo{year}{1982}).

\bibitem{fawcett2006introduction}
\bibinfo{author}{Fawcett, T.}
\newblock \bibinfo{journal}{\bibinfo{title}{An introduction to roc analysis}}.
\newblock {\emph{\JournalTitle{Pattern recognition letters}}}
  \textbf{\bibinfo{volume}{27}}, \bibinfo{pages}{861--874}
  (\bibinfo{year}{2006}).

\bibitem{bengio2003neural}
\bibinfo{author}{Bengio, Y.}, \bibinfo{author}{Ducharme, R.},
  \bibinfo{author}{Vincent, P.} \& \bibinfo{author}{Jauvin, C.}
\newblock \bibinfo{journal}{\bibinfo{title}{A neural probabilistic language
  model}}.
\newblock {\emph{\JournalTitle{Journal of machine learning research}}}
  \textbf{\bibinfo{volume}{3}}, \bibinfo{pages}{1137--1155}
  (\bibinfo{year}{2003}).

\bibitem{wieting2015towards}
\bibinfo{author}{Wieting, J.}, \bibinfo{author}{Bansal, M.},
  \bibinfo{author}{Gimpel, K.} \& \bibinfo{author}{Livescu, K.}
\newblock \bibinfo{journal}{\bibinfo{title}{Towards universal paraphrastic
  sentence embeddings}}.
\newblock {\emph{\JournalTitle{arXiv preprint arXiv:1511.08198}}}
  (\bibinfo{year}{2015}).

\bibitem{japkowicz2002class}
\bibinfo{author}{Japkowicz, N.} \& \bibinfo{author}{Stephen, S.}
\newblock \bibinfo{journal}{\bibinfo{title}{The class imbalance problem: A
  systematic study}}.
\newblock {\emph{\JournalTitle{Intelligent data analysis}}}
  \textbf{\bibinfo{volume}{6}}, \bibinfo{pages}{429--449}
  (\bibinfo{year}{2002}).

\bibitem{world1992international}
\bibinfo{author}{Organization, W.~H.} \emph{et~al.}
\newblock \bibinfo{journal}{\bibinfo{title}{International statistical
  classification of diseases and health related problems, 10th revision}}.
\newblock {\emph{\JournalTitle{Geneva: WHO}}}  (\bibinfo{year}{1992}).

\end{thebibliography}

\section*{Acknowledgements}
The authors thank Devendra Singh Sachan for his sharing of pre-trained word vectors.

\section*{Author contributions statement}
H.S. and P.X. conceived and designed the study. H.S. processed the data and performed the experiments. H.S., P.X., Z.H. wrote the paper. M.Z. and E.P.X. take responsibility for the paper as co-senior authors. All authors reviewed the manuscript. 

\section*{Additional information}
\subsection*{Supplementary information}
The supplementary material is available on request.

\subsection*{Competing interests}
The authors declare that they have no competing interests.

\end{document}